\pdfoutput=1

\documentclass[11pt,a4paper]{article}

\usepackage{times,latexsym}
\usepackage{url}
\usepackage[T1]{fontenc}
\usepackage{paralist}
\usepackage[dvipsnames]{xcolor}

\usepackage[acceptedWithA]{tacl2021v1}

\usepackage{xspace,mfirstuc,tabulary}

\newcommand{\newmaterial}[1]{#1}

\newif\iftaclinstructions
\taclinstructionsfalse 
\iftaclinstructions
\renewcommand{\confidential}{}
\renewcommand{\anonsubtext}{(No author info supplied here, for consistency with
TACL-submission anonymization requirements)}
\newcommand{\instr}
\fi

\iftaclpubformat 

\else

\fi

\usepackage{booktabs}
\usepackage{multirow}
\usepackage{graphicx}
\usepackage{hyperref}
\newcommand{\metric}[1]{\emph{#1}}
\newcommand{\citeinp}[2]{\metric{#1}:~\citealt{#2}}
\newcommand{\myparagraph}[1]{\paragraph{#1}}
\newcommand{\name}[0]{\textsc{EnsembEval}}
\newcommand{\datasetname}[1]{#1}

\usepackage{wrapfig}

\title{Surveying the Landscape of Image Captioning Evaluation:\\ A Comprehensive Taxonomy, Trends and Metrics Analysis}

\author{
  Uri Berger$^\diamond\dagger$ 
  \and
  Gabriel Stanovsky$^\diamond$
  \and
  Omri Abend$^\diamond$
  \and
  Lea Frermann$^\dagger$\vspace{1em}
  \\
  $^\diamond$School of Computer Science and Engineering, The Hebrew University of Jerusalem
  \\
    \texttt{\{uri.berger2, gabriel.stanovsky, omri.abend\}@mail.huji.ac.il}
    \\
  $^\dagger$School of Computing and Information Systems, University of Melbourne
  \\
  \texttt{lea.frermann@unimelb.edu.au}
}

\date{}

\begin{document}
\maketitle
\begin{abstract}
The task of image captioning has recently been gaining popularity, and with it the complex task of evaluating the quality of image captioning models. 
In this work, we present the first survey and taxonomy of over 70 different image captioning metrics and their usage in hundreds of papers, specifically designed to help users select the most suitable metric for their needs.
We find that despite the diversity of proposed metrics, the vast majority of studies rely on only five popular metrics, which we show to be weakly correlated with human ratings.
We hypothesize that combining a diverse set of metrics can enhance correlation with human ratings. As an initial step, we demonstrate that a linear regression-based ensemble method, which we call \name{}, trained on one human ratings dataset, achieves improved correlation across five additional datasets,
showing there is a lot of room for improvement by leveraging a diverse set of metrics.\footnote{Our code and data are available at \\
\iftaclpubformat
\href{https://github.com/uriberger/caption_evaluation}{github.com/uriberger/caption\_evaluation}.
\else
\href{https://anonymous.4open.science/r/caption_evaluation-6021/}{anonymous.4open.science/r/caption\_evaluation-6021}.
\fi
}

\end{abstract}

\section{Introduction}
\begin{figure}[tb]
    \centering
    \includegraphics[width=\columnwidth]{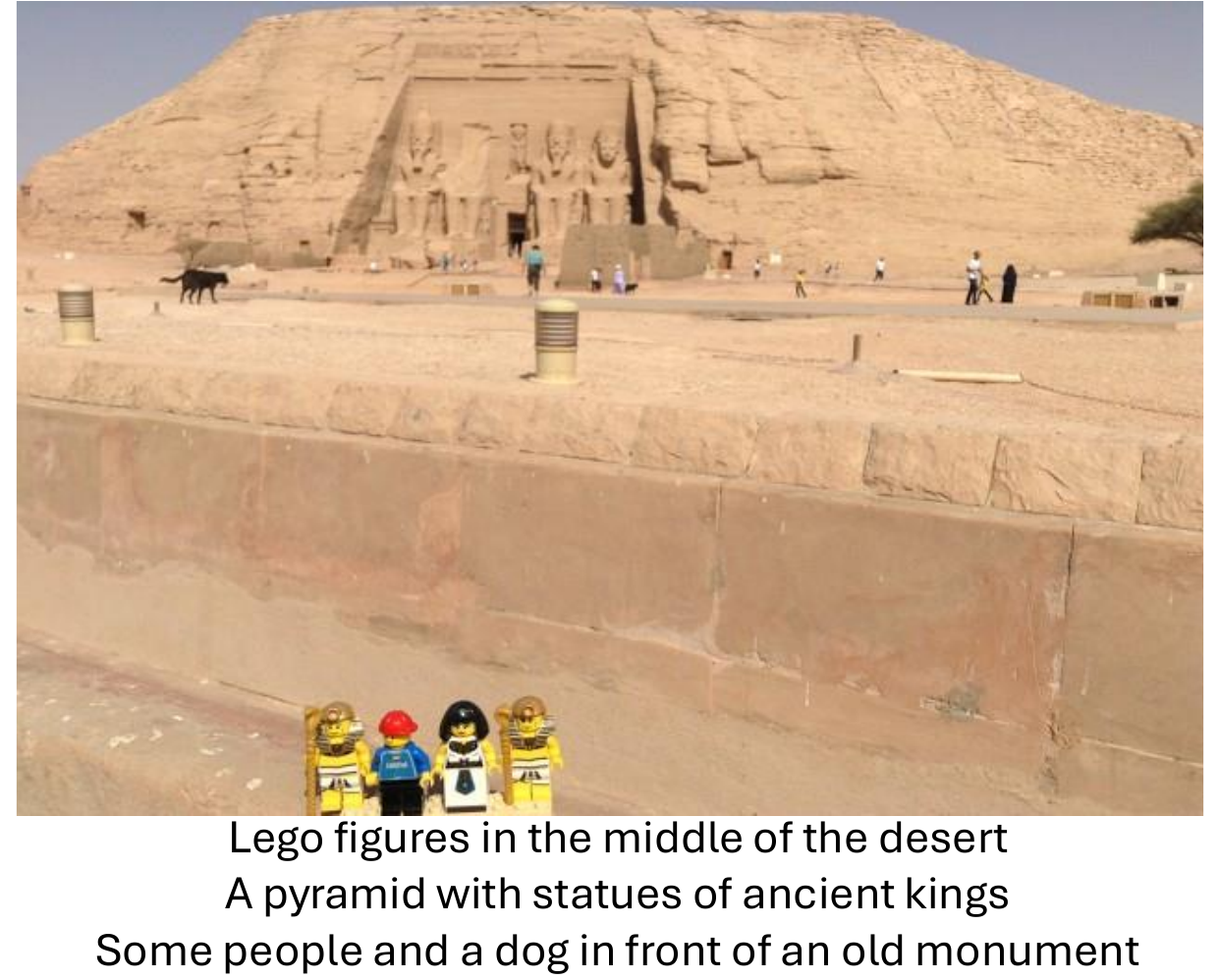}
    \caption{An image with various valid captions that have no overlap in content words, exemplifying some of the challenges in evaluating image captioning \citep[taken from Crossmodal-3600;][]{thapliyal-etal-2022-crossmodal}.} 
    \label{fig:image_example}
\end{figure}

\begin{figure}[tb]
    \centering
    \includegraphics[width=\columnwidth]{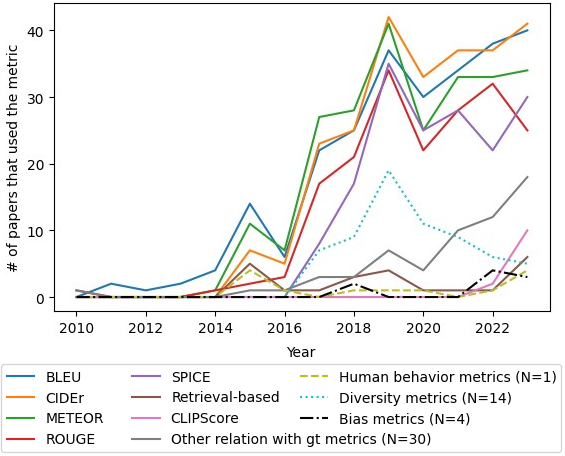}
    \caption{Metrics usage over the years.
    Metrics used in $\le 1$ papers are omitted. $N=$ indicates number of metrics in this category.
    }
    \label{fig:metric_usage}
\end{figure}

Evaluating the output of image captioning is challenging for several reasons. 
First, as in other generative tasks, multiple outputs may be valid for the same input, as illustrated in Figure~\ref{fig:image_example}, where different valid captions for the same image have no overlap in content words.
Second, it involves bridging across modalities, requiring evaluators to compare text and images, unlike most generative tasks that only involve textual information.
Attesting to the difficulty and importance of image captioning is the sheer volume of  metrics proposed for this task, and their usage in hundreds of image captioning models (Figure~\ref{fig:metric_usage}).

In this work, we survey the different approaches proposed for evaluating image captioning, and provide a first taxonomy comprising over 70 metrics, which were largely developed independently of one another.
We examine usage patterns of metrics over the past 14 years and demonstrate that combining multiple metrics enhances correlation with human ratings.

We begin by exploring various definitions of image captioning presented in previous studies and examining their impact on evaluation (Section~\ref{sec:task_definition}). Then,
in Section~\ref{sec:methods}, we systematically  examine all relevant  papers from 15 major venues in NLP, vision and machine learning between 2010 and 2024. This approach yields a body of work consisting of 71 different automatic evaluation metrics and 5 human evaluation paradigms, used in over 300 papers. 

Next, we organize both automatic metrics and human evaluation in a principled taxonomy~(Section~\ref{sec:taxonomy}). Our taxonomy is the first to comprehensively cover all automatic metrics used in image captioning research.
We design the taxonomy dimensions with the needs of metric users in mind, focusing on dimensions such as the candidate's property each metric aims to measure.
We also propose the first taxonomy for human evaluation metrics, categorized into groups such as comparative evaluation and scale rating evaluation.

We then survey the use of evaluation methods  across 314 papers~(Section~\ref{sec:usage_analysis}). 
Interestingly, we find that despite the wealth of different metrics proposed for the task, the vast majority of examined papers use only five simple metrics (BLEU, METEOR, ROUGE, CIDEr, SPICE),
although there has been a recent increase in the adoption of alternative metrics.
Furthermore, we find that the use of human evaluation is declining in recent years and that significance and inter-annotator agreement are rarely reported.

Finally, in a series of evaluations, we show that the five most popular metrics show only a weak correlation with human ratings, whereas lesser-known metrics show a much higher correlation. This underscores the value of a systematic and comprehensive survey in identifying these lesser-known but more effective metrics.
We then show that an ensemble of selected metrics we name \name{}, optimized for diversity using a feature selection algorithm, achieves enhanced correlation with human ratings. This paves the way for future research on metrics that consider multiple aspects simultaneously.

To recap, our contributions are threefold. First, we introduce the first comprehensive taxonomy of automatic captioning evaluation metrics categorized by user-oriented aspects, helping future model developers select the most suitable metric based on the specific property they want their model to excel in.
Second, we examine key trends in image captioning evaluation, highlighting a shift from reliance on five simple metrics toward alternative approaches.
Finally, we demonstrate that a simple linear regression-based ensemble method improves correlation with human ratings and provide the code publicly for future use.

\section{Task Definition} \label{sec:task_definition}

Image captioning can be motivated by diverse goals, and what counts as a `good' caption heavily depends on the system's purpose and context. Accurate evaluation requires explicitly defining a system's purpose, a step often overlooked or only implicitly addressed in previous work. We start with a review of common definitions.

Early research framed image captioning as a means to assess vision-and-language models based on image-caption semantics, rather than practical applications~\cite{hodosh2013framing}. 
More recently, the task has often been defined simply as describing the visual content of an image or scene in natural language~\cite{stefanini2022show, Nallapaneni_Konakanchi_2023, gonzalez2024metrics}. Yet, this definition does not specify what visual content should be described or the intended nature of the description. These aspects depend on both the \textit{purpose} and the \textit{context} of the description, which we now discuss in detail.

Descriptions serve different \textbf{purposes}, each with distinct requirements. For example, captions designed for accessibility should \textit{replace} the image, while captions in news articles typically \textit{augment} or \textit{link} the content of the image and article~\cite{kreiss-etal-2022-concadia}. 
In practice, most studies focus on image \textit{replacement}, i.e., generating descriptions intended to serve as substitutes for the image, as defined by \citet{kreiss-etal-2022-concadia}.

\newmaterial{
The exact purpose of image captioning is often implicit in image captioning papers.
\citet{hodosh2013framing} discusses the task in detail, and adopts a number of fundamental
theoretical distinctions as to the purposes of captioning, taken from library science \citep{shatford1986analyzing,jaimes1999conceptual}.
Specifically, they argue that captions that focus on conceptual descriptions (rather than low-level ``perceptual'' descriptions) are the most relevant for image understanding,  the most commonly tackled purpose in NLP. Within this category, an important sub-distinction is between specific captions, that identify entities by their names (e.g., Great Pyramid of Giza), and generic ones (e.g., that instead identify an object as a pyramid). Image understanding usually (albeit, not universally) targets the latter. 
}


The second key factor is the \textbf{context} in which a caption is generated. For instance, the top caption in Figure~\ref{fig:image_example} may suit Lego's Instagram account, while the middle caption might be more appropriate for a travel guide.



While our survey covers all works on captioning evaluation --- irrespective of purpose or context --- our experiments on metric performance and ensembling (Section~\ref{sec:experiments}) focus on the {\it replacement} task, specifically.

\section{Literature Survey} \label{sec:methods}
We perform a systematic review and gather usage statistics of metrics in conference papers on English image captioning.
Overall, we examined 314 papers, and identified 71 distinct metrics.

\begin{figure}[tb]
    \centering  
    \includegraphics[width=\columnwidth]{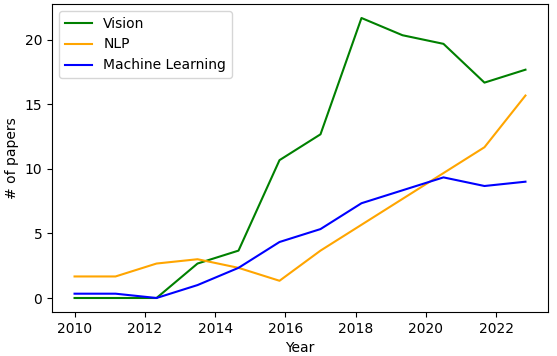}
    \caption{Annual number of image captioning papers per community, smoothed by convolving with a window size of 3 years.}
    \label{fig:papers_per_community}
\end{figure}

\myparagraph{Scope.}
We focus on a large set of 15 venues spanning three distinct research communities (Table~\ref{tab:venues}). Figure~\ref{fig:papers_per_community} shows the annual number of papers for each community. We begin collecting data starting from 2024 backwards until reaching the first year with no relevant papers, which is 2009.

\myparagraph{Search strategy.}
To identify image captioning related papers within a venue, we search the venue proceedings for papers with the substrings \emph{caption} or \emph{description} in their titles, assuming relevant titles are likely to include variations of the phrases \emph{image captioning} or \emph{image descriptions}.

\myparagraph{Filtering.}
Next, we manually filter out papers that we deem irrelevant to our review. Specifically, we exclude: 1) papers on other types of captioning such as video captioning; 2) papers unrelated to captioning that happen to have \emph{caption} or \emph{description} in their titles; 3) papers that focus on languages other than English; and 4) papers that did not conduct any experiments (e.g., review papers).

\myparagraph{Collected data.}
We collect two types of information.
First, we intend to describe the types of metrics used in previous work. For each metric introduced or used in the papers we examined, we document its name, implementation details and the paper in which it was introduced. To the best of our knowledge, we are the first to systematically collect such a comprehensive set of metrics, as previous research has focused on a few well-known metrics (see Section~\ref{sec:related_work}).

Second, we seek to analyze the patterns of metrics usage. Therefore, we record which metrics were used in each examined paper.\footnote{We only document metrics that were used in the main body of the paper, omitting metrics used in the Appendix.}

\begin{table}[t]
\centering
\begin{small}
\begin{tabular}{cl}
\toprule
\textbf{Community} & \textbf{Conferences} \\
\midrule
\multirow{2}{*}{NLP} & AACL, ACL, CoNLL, EACL \\
 & EMNLP, NAACL, TACL, *SEM \\
CV & CVPR, ICCV, ECCV \\
ML & Neurips, ICML, ICLR, AAAI \\
\bottomrule
    \end{tabular}
    \end{small}
    \caption{Venues included in our review. NLP: Natural Language Processing. CV: Computer Vision. ML: Machine Learning.}
    \label{tab:venues}
\end{table}

\section{Taxonomy} \label{sec:taxonomy}
Here, we introduce a taxonomy of the metrics from the systematic review outlined in Section~\ref{sec:methods}.

This work is novel in two respects: we are the first to comprehensively include all metrics proposed and used in previous studies, rather than focusing on a limited set of well-known metrics. Second, we are the first to describe a taxonomy for human evaluation methods.

\subsection{Automatic Evaluation} \label{sec:automatic_evaluation}

\definecolor{lightblue}{rgb}{0.8118, 0.8941, 1}
\definecolor{lightred}{rgb}{1, 0.851, 0.851}

\begin{figure*}[tb]
    \centering
    \includegraphics[width=\textwidth]{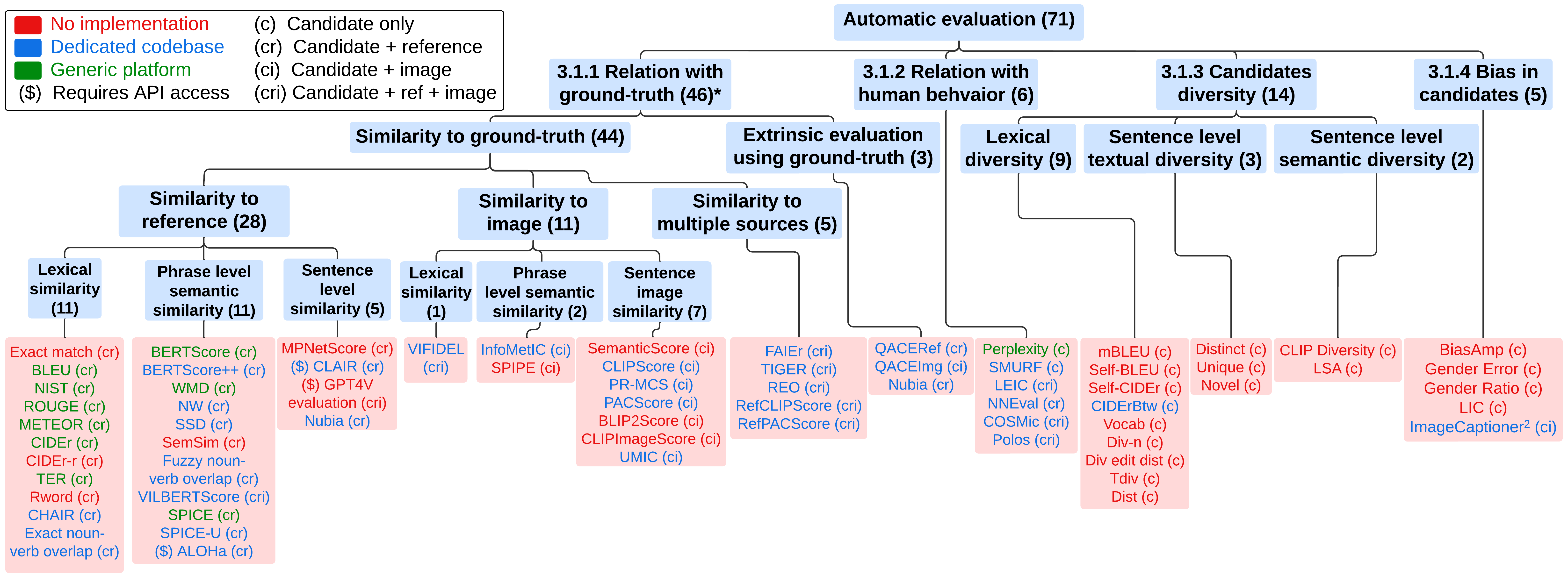}
    \caption{A taxonomy of automatic metrics for image captioning evaluation. For each category, we list the number of metrics in parentheses. Top categories are preceded by the relevant section number. \textcolor{lightblue}{Blue} rectangles contain categories, \textcolor{lightred}{red} contain metrics. * -- Nubia is assigned to two distinct categories under \emph{Relation with ground-truth}.}
    \label{fig:taxonomy}
\end{figure*}


\paragraph{Motivation.}
Given the large number of 71 automatic evaluation methods identified in our survey, it is a challenge to stay updated on the latest metrics. As a result, the common practice is to rely on previously used metrics, typically the five dominant ones -- BLEU, CIDEr, METEOR, ROUGE, and SPICE -- as shown in Figure~\ref{fig:metric_usage}.

To address this issue, we propose a taxonomy that helps users select the most suitable metric. We categorize metrics based on the specific property they assess in the candidate caption.

\paragraph{Method.}
We begin by categorizing the metrics based on the evaluated property. For example, BLEU, BERTScore and CLIPScore are classified under \textit{lexical similarity to reference}, \textit{phrase level semantic similarity to reference} and \textit{semantic similarity to image}, respectively.\footnote{One metric (\textit{Yngve score}~\citet{Liu_2019_ICCV}) did not align with any category, another (\textit{Nubia}~\citet{kane-etal-2020-nubia}) fit into two categories.}

Next, we incrementally group together pairs of categories based on their similarity to one another. For instance, \textit{lexical similarity to reference} and \textit{phrase level semantic similarity to reference} are both grouped under \textit{similarity to reference}, while \textit{semantic similarity to image} falls under \textit{similarity to image}.
We repeat this process iteratively until only one overarching category remains.

Finally, following the process of~\citet{nickerson2013method}, we extend the taxonomy with three additional dimensions, again motivated by the overarching goal to support users in their choice of metrics:

\begin{compactitem}
    \item \textbf{Input type:} The data provided to the metric (candidate only, candidate + reference, candidate + image, candidate + reference + image).
    \item \textbf{Implementation availability:} Availability of the metric's implementation (none, dedicated codebase, or integration within a generic platform like HuggingFace).
    \item \textbf{API access requirements:} Whether the metric relies on a closed model API access.
\end{compactitem}
The initial taxonomy was constructed by the first author, followed by validation from the remaining authors. Any disagreements were resolved through majority voting.
The final taxonomy is visually presented in Figure~\ref{fig:taxonomy}. In the following sections, we describe the various automatic evaluation methods.

\subsubsection{Relation with Ground-Truth}
The largest and most widely used class of automatic methods use a ground-truth source (either a reference caption or the image) as a basis for evaluating the candidate.
We categorize these methods based on the relation they examine (similarity/extrinsic) and the nature of the ground-truth  (reference, image, both) they use.

\myparagraph{Similarity to reference.}

Early automatic methods, still the most common today, compare the candidate caption to a reference caption.

\emph{Lexical similarity:}
Some metrics compare the candidate and reference captions at the word level using a straightforward textual comparison,
most na\"ively using exact match~\cite{Kang_2023_ICCV}.
More advanced methods include computing $n$-gram overlap (\citeinp{BLEU}{papineni-etal-2002-bleu}, \citeinp{NIST}{doddington2002automatic}, \citeinp{ROUGE}{lin-2004-rouge}, \citeinp{METEOR}{banerjee-lavie-2005-meteor}), $n$-gram TF-IDF (\citeinp{CIDEr}{vedantam2015cider}, \citeinp{CIDEr-r}{oliveira-dos-santos-etal-2021-cider}), and calculating minimal edits to match the reference~\citep[\metric{TER},][]{snover-etal-2006-study}. Other metrics include the number of reference words in the candidate~\citep[\metric{Rword},][]{cho-etal-2022-fine} and semantic comparisons including synonyms~\citep[\metric{CHAIR},][]{rohrbach-etal-2018-object}.
Finally, some studies measure the precision and recall of specific word categories, including parts of speech~\citep[e.g., \metric{Exact noun/verb overlap}][]{chan-etal-2023-ic3}, objects~\cite{wang-etal-2021-ecol} and named entities~\citep[common in newsimage captioning, e.g.,][]{zhang-wan-2023-exploring}.

\emph{Phrase-level semantic similarity:}
Others compare the semantics of sentence elements, most commonly using phrase embeddings, either with context (\citeinp{BERTScore}{zhang2019bertscore}, \citeinp{BERTScore++}{yi-etal-2020-improving}) or without (\citeinp{WMD}{pmlr-v37-kusnerb15}, \citeinp{NW}{cornia2019show}, \citeinp{SSD}{takmaz-etal-2020-generating}, \citeinp{SemSim}{nag-chowdhury-etal-2021-exploiting}, \citeinp{Fuzzy noun/verb overlap}{chan-etal-2023-ic3}).
\metric{VILBERTScore}~\cite{lee-etal-2020-vilbertscore} enriches phrase embeddings by injecting visual information.
The SPICE family (\citeinp{SPICE}{anderson2016spice}, \citeinp{SPICE-U}{wang2020towards}) compares the components of scene graphs of the candidate and reference captions.
\metric{ALOHa}~\cite{petryk-etal-2024-aloha} prompts Large Language Models (LLMs) to identify object phrases in both candidate and references and computes the similarity of these phrases' embeddings.

\emph{Sentence-level similarity:}
Some metrics compare semantics at the sentence level.
\metric{MPNetScore}~\cite{black2024vixen} compares sentence level embeddings.
Others (\citeinp{CLAIR}{chan-etal-2023-clair}, \citeinp{GPT4V evaluation}{ge2024visual}) prompt LLMs to compare the candidate and reference caption.
\metric{Nubia}~\cite{kane-etal-2020-nubia} explicitly trains a model to evaluate sentence similarity.

\emph{Extracted information similarity:}
In specific domains (e.g., medical image captioning), it is common to extract relevant information from both the reference and candidate captions for comparison~\citep[e.g., a list of diseases from the captions in clinical image captioning,][]{nishino-etal-2022-factual}.

\myparagraph{Similarity to image.}
Recent work suggests comparing the candidate caption to the image instead of a reference caption. These metrics are also known as reference-free metrics.

\emph{Lexical similarity:}
\metric{VIFIDEL}~\cite{madhyastha-etal-2019-vifidel} compares object phrases in the candidate with names of objects identified in the image.

\emph{Phrase-level semantic similarity:}
\metric{InfoMetIC}~\cite{hu-etal-2023-infometic} compares text and visual token embeddings.
\metric{SPIPE}~\cite{NEURIPS2022_6e4df340} is a variation of SPICE where the scene graphs of the candidate and image are compared.

\emph{Sentence-image similarity:}
Several recent works  embeded sentences and images in a joint visual-textual space and computed the similarity of candidate-image embeddings (\citeinp{Semantic Score}{Dognin_2019_CVPR}, \citeinp{CLIPScore}{hessel-etal-2021-clipscore}, \citeinp{PR-MCS}{kim-etal-2023-pr}, \citeinp{PACScore}{Sarto_2023_CVPR}, \citeinp{BLIP2Score}{zeng2024meacap}).
\metric{CLIPImageScore}~\cite{ge2024visual} use text-to-image methods to generate an image based on the candidate and computes the embedding similarity between this generated image and the original image.
\metric{UMIC}~\cite{lee-etal-2021-umic} explicitly trains models to predict similarity of candidate caption and image.

\emph{Retrieval-based methods:}
A popular evaluation protocol involves using text-to-image retrieval on the test set~\citep[e.g.,][]{Kornblith_2023_ICCV}. In this setting, a candidate caption is matched with each image in the test set using an image-text matching model (most commonly CLIP), and the images are ranked by their matching score with the candidate. Then, recall@n is applied as the evaluation metric, with typical values for $n$ being 1, 5, and 10.

\myparagraph{Similarity to multiple sources.}
Some methods use both the image and reference captions as sources for comparison.
\metric{FAIEr}~\cite{Wang_2021_CVPR} compares the candidate scene graph with a fusion of the image and reference scene graphs.
\metric{TIGER}~\cite{jiang-etal-2019-tiger} and \metric{REO}~\cite{jiang-etal-2019-reo} compare the candidate-image joint embedding vector with the reference-image joint embedding vector.
\metric{RefCLIPScore}~\cite{hessel-etal-2021-clipscore} and \metric{RefPACScore}~\cite{Sarto_2023_CVPR} compute the harmonic mean of candidate-image similarity with candidate-reference similarity.

\myparagraph{Extrinsic evaluation using ground-truth.}
Several NLP tasks focus on understanding the relation between two input sources. Some metrics use models trained for this task with the candidate and a ground-truth source as the inputs.

\emph{Question generation and question answering.}
\citet{lee-etal-2021-qace-asking} generate questions from the candidate and score it by how well a question-answering model answers using the reference caption (\metric{QACERef}) or the image (\metric{QACEImg}).

\emph{Natural language Inference (NLI).}
Nubia~\cite{kane-etal-2020-nubia} use NLI models with the reference as the premise and the candidate as the hypothesis.

\subsubsection{Relation With Human Behavior}

The primary aim of captioning systems developed in recent years is to imitate human behavior, both in the text they generate and in the evaluation process. Metrics in this category strive to explicitly assess these properties.

\myparagraph{Candidate fluency.}
Some works~\citep[e.g.,][]{ou-etal-2023-pragmatic} evaluate the candidate's \metric{perplexity} by employing an off-the-shelf large language model \citep[commonly GPT-2,][]{radford2019language} to compute the probability of the candidate's tokens.
\metric{SMURF}~\cite{feinglass-yang-2021-smurf} computes the activation flow through a Transformer model self-attention layers, assumed to increase when the candidate differs greatly from typical captions.

\myparagraph{Is the candidate human-like?}
\metric{LEIC}~\cite{Cui_2018_CVPR} and \metric{NNEval}~\cite{Sharif_2018_ECCV} measure how likely the candidate is to deceive a model that discriminates between human-generated and machine-generated sentences.

\myparagraph{Human rating prediction.}
\metric{COSMic}~\cite{inan-etal-2021-cosmic-coherence} and \metric{Polos}~\cite{Wada_2024_CVPR} explicitly train a model to predict human scores.

\subsubsection{Candidate Diversity}

Several methods measure the diversity of generated captions. All assume a set of generated captions $S$ for which diversity is measured, where $S$ can either consist of candidates for the entire test set, a set of similar images or a single image.

\emph{Lexical diversity:}
While some works use lexical similarity methods such as BLEU (\citeinp{mBLEU}{Shetty_2017_ICCV}, \citeinp{self-BLEU}{10.1145/3209978.3210080}) and CIDEr (\citeinp{Self-CIDEr}{Wang_2019_CVPR}, \citeinp{CIDErBtw}{wang2020compare}) on pairs of captions from $S$,
others specifically designed methods to measure lexical diversity in $S$ (\citeinp{Vocab}{Shetty_2017_ICCV}, \citeinp{Div-$n$}{Shetty_2017_ICCV}, \citeinp{diversity edit distance}{NEURIPS2018_8bf1211f}, \citeinp{Tdiv}{Liu_2019_ICCV}, \citeinp{Dist}{Liu_2019_ICCV}).

\emph{Sentence-level textual diversity:}
Other methods compute the percentage of distinct candidates in $S$~\citep[\metric{Distinct}, AKA \metric{Unique},][]{NIPS2017_4b21cf96}, or the percentage of sentences not seen in the training set~\citep[\metric{Novel},][]{NIPS2017_4b21cf96}.

\emph{Sentence-level semantic diversity:}
Recent methods propose to compare the semantics of candidates as an indicator for diversity.
\metric{LSA}~\cite{Wang_2019_CVPR} computes a matrix $K$ where $K_{i,j}$ is the dot-product between the bag-of-words vectors of captions $i,j$ in $S$, and calculates the singular vector decomposition (SVD) of $K$.
\metric{CLIP diversity}~\cite{Li_2023_ICCV} computes the similarity of the CLIP embeddings of captions in $S$.

\subsubsection{Bias in Candidates}
Various metrics have been proposed to quantify the extent of bias evident in captioning models.
\metric{BiasAmp}~\cite{zhao-etal-2017-men} measures the amplification of bias by the model compared to the training set by comparing the correlation between predefined attributes and activities (e.g., female-cooking) in model- and human-generated captions.
\citet{Hendricks_2018_ECCV} introduce two metrics: \metric{Gender error}, which assumes that images are labeled as male or female and calculates the number of gender misclassified words in the candidates, and \metric{Gender ratio}, which defines male or female sentences based on the inclusion of predefined gender-related words and computes the ratio of male candidates to female candidates.
Others measure bias amplification by training models to predict protected attributes given a caption~\citep[\metric{LIC},][]{Hirota_2022_CVPR} or an image-caption pair~\citep[\metric{ImageCaptioner$^2$},][]{abdelrahman2024imagecaptioner2}, and computing the difference between accuracies when training on candidates versus references.

\subsection{Human Evaluation}

While automatic evaluation offers clear benefits in terms of scale and consistency, it still serves as a surrogate for human evaluation. To ensure that the improvements shown by automatic methods are genuine, many image captioning studies also apply human evaluation methods to a subset of the data.

The human evaluation taxonomy development process followed a similar approach to that in automatic evaluation (Section~\ref{sec:automatic_evaluation}). In this case, a single dimension emerged: the evaluation framework.

\myparagraph{Scale rating.}
Several studies direct human participants to evaluate candidates on a discrete scale based on various attributes, such as relevance~\cite{maeda-etal-2023-query}, fluency~\cite{wu-etal-2023-cross2stra} and descriptiveness~\cite{NEURIPS2023_fa1cfe4e}.

\myparagraph{Comparative.}
A different approach presents pairs of captions to human participants and asks them to decide which one is better without knowing their source. Early studies compared candidates to references~\cite{kuznetsova-etal-2014-treetalk, yatskar-etal-2014-see}, while recent research compares candidates with captions generated by baseline models~\cite{tanaka-etal-2024-content, ge2024visual}.

\myparagraph{Yes/no questions.}
Certain studies involve human participants answering yes/no questions such as whether the candidate exhibits human-like qualities~\cite{Yao_2019_ICCV} or describes all the objects in the image~\cite{Chen_2022_CVPR}.

\myparagraph{Retrieval.}
Another popular approach is to ask participants to perform text-to-image retrieval given the candidate caption~\cite{Wang_2019_CVPR, ou-etal-2023-pragmatic}.

\myparagraph{Answer questions using the candidate.}
\citet{nie-etal-2020-pragmatic} prompts participants with questions and requires them to answer using the candidate provided, without showing the images.

\section{Metrics Usage Analysis} \label{sec:usage_analysis}
In this section, we analyze the data from Section~\ref{sec:methods} to identify usage patterns and present the main findings, excluding 2024 as the data was collected before the year ended.
Appendix~\ref{sec:app_metric_usage} lists metric usage for all 314 reviewed papers.

\subsection{Automatic Evaluation} \label{sec:automatic_evaluation_trends}

\myparagraph{A set of 5 metrics dominates.} 
Figure~\ref{fig:metric_usage} displays the number of papers using different metrics each year. Given the large number of metrics identified (71), we only plot usage for the seven most common metrics, clustering all other metrics by the categories from Section~\ref{sec:taxonomy}.

\begin{figure}
    \centering
    \includegraphics[width=\columnwidth]{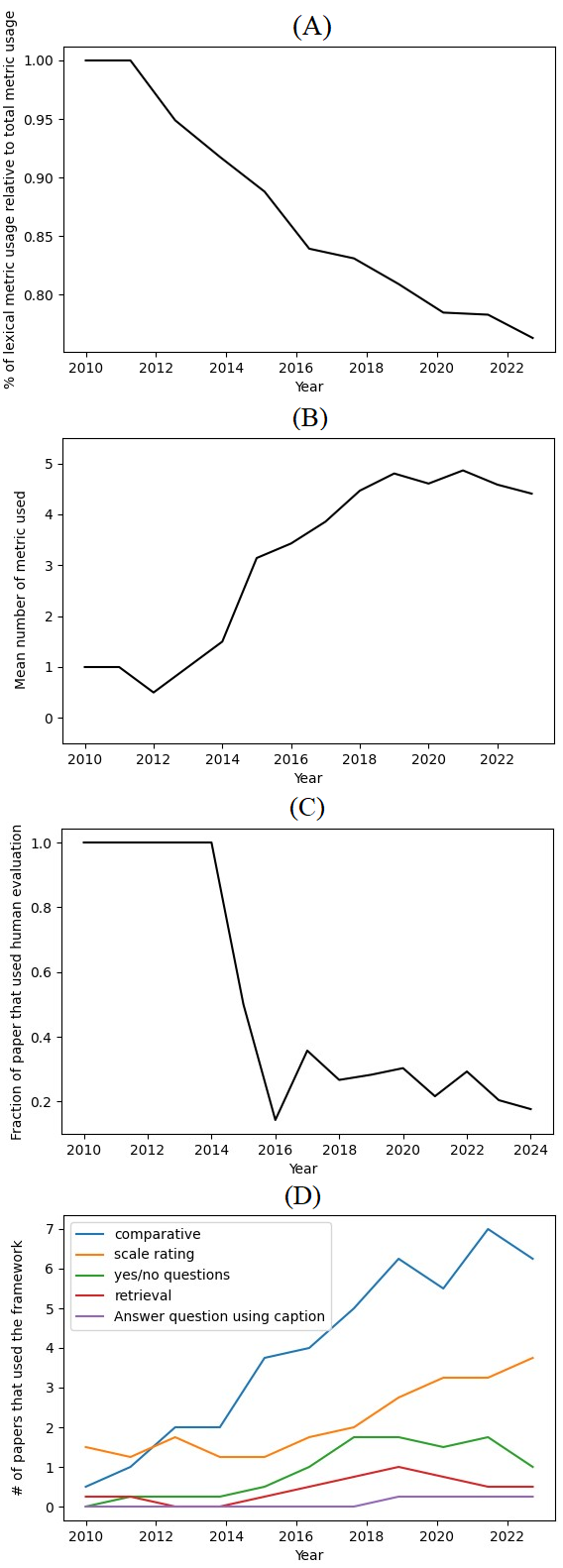}
    \caption{
    Metrics usage analysis plots. (A): Percentage of lexical metric usage relative to total metric usage each year, smoothed by convolving with a window size of four years. (B): Mean number of metrics used per paper each year. (C): Fraction of papers that performed human evaluation each year. (D): Human evaluation frameworks usage over the years, smoothed by convolving with a window size of four years.
    }
    \label{fig:analysis_plots}
\end{figure}

The figure shows that since 2015, five metrics -- BLEU, CIDEr, METEOR, ROUGE, and SPICE (henceforth, the five dominant metrics) -- have been used substantially more frequently than all other metrics.
This trend has begun to shift in recent years, with an increase in the use of other metrics.
Notably, four of the dominant metrics (all except SPICE) are lexical similarity metrics, criticized for failing to capture semantic similarity~\cite{gimenez-marquez-2007-linguistic}. Figure~\ref{fig:analysis_plots} (A) illustrates a decline in the usage of lexical metrics in recent years relative to overall metric usage, excluding diversity and bias metrics from this count.

\myparagraph{The number of metrics per paper has settled at 4-5.}
Figure~\ref{fig:analysis_plots}
(B)
shows a gradual increase in the mean number of metrics used per paper each year until reaching approximately 4.5 in 2019, and remaining in the range of 4-5 since then.

\subsection{Human Evaluation}

\myparagraph{Human evaluation usage is decreasing.}
Figure~\ref{fig:analysis_plots}
(C)
shows the fraction of papers with human evaluation each year. From 2010 to 2014, all papers conducted human evaluation. However, the publication of large image-caption datasets made this more challenging, and beginning at 2015 (a year after MSCOCO's release) we see a general decline in the use of human evaluation, likely due to improvements in automatic evaluation metrics.

\myparagraph{Comparative is the most common human framework.}
Figure~\ref{fig:analysis_plots}
(D)
presents the number of papers using each human evaluation framework per year. The comparative framework is most frequently used, closely followed by scale rating.

\myparagraph{Significance and agreement are rarely reported.}
Of 94 papers that used human evaluation (29.9\% of the papers we documented), only 19 reported significance, which is crucial in human evaluation because, due to the high cost of human labor, only a small subset of the test set is assessed. Moreover, results may be unreliable if annotator's agreement is low; however, only 6 of these 94 papers report annotator agreement.

\section{Experiments} \label{sec:experiments}
Prior captioning evaluation work assumed an automatic metric's quality depends on its alignment with human evaluations. To facilitate this, human rating datasets were created, where participants rated candidate captions. A common approach~\citep[among others,][]{anderson2016spice, hessel-etal-2021-clipscore} is to evaluate metrics by applying them to the candidates in these datasets and measuring the correlation of their scores with the human ratings.
We follow this convention but extend prior research by incorporating metrics that have not been previously studied in our experiments.

Throughout our experiments, we focus on the \textit{replacement} subtask (see Section~\ref{sec:task_definition}). This definition -- producing descriptions intended to serve as substitutes for the image -- aligns most closely with the human rating datasets, where annotators are instructed to evaluate whether the text accurately reflects the image's content without adding information not present in the image.\footnote{See Appendix \ref{sec:app_datasets} for annotators instructions for existing human rating datasets.}

\subsection{Experimental Setup}

\myparagraph{Datasets.}
We experiment with the following 7 human rating datasets.

\datasetname{Flickr8k-Expert} and \datasetname{Flickr8k-CF}~\cite{hodosh2013framing} include human ratings for captions on the 1000 images of the Flickr8k test set. 
The captions are human-generated captions selected from the test set, where each image is associated with a set of captions (not necessarily initially generated for that image).
Flickr8k-Expert includes 5,822 captions rated by a small and controlled group of native English speakers on a scale from 1 to 4.
Flickr8k-CF comprises 47,830 captions, each rated by three or more crowd-sourced workers using a binary label (describes/doesn't describe the image). For each caption, we take the percentage of positive ratings as the final ground truth score.

\datasetname{Composite}~\cite{aditya2015images} includes 11,985 human ratings for captions of 3,925 images from Flickr8k, Flickr30k and MSCOCO. Each image is paired with two model-generated and one original reference captions. Each caption is rated for correctness and thoroughness on a scale from 1 to 5. Following previous studies~\cite{anderson2016spice}, we use the correctness rating (also termed ``relevance'' in some prior research) as the score.

\datasetname{THumB}~\cite{kasai-etal-2022-transparent} features human ratings of captions for 500 images sourced from MSCOCO.
Each image is associated with four model-generated and one original reference captions.
Annotators provide ratings for precision and recall (on a scale of 1 to 5), deducting some penalty points from their average in case of lack of fluency, conciseness and incluse language to reach the overall score. We take this overall score as the score for this caption. 

\datasetname{Polaris}~\cite{Wada_2024_CVPR} contains 131,020 human ratings for captions generated for 13,691 images from MSCOCO and nocaps~\cite{Agrawal_2019_ICCV}. Ratings were provided on a five-point scale and then transformed to values in the range [0,1] using min-max normalization.
The captions were generated by 10 different models. The dataset is partitioned into training, validation and test sets.

\datasetname{Pascal50S}~\cite{vedantam2015cider} contains 4000 pairs of candidate captions for images from the Pascal Sentence dataset~\cite{rashtchian-etal-2010-collecting}, evenly distributed across four categories: both captions are human-generated for the image (HC), both are human-generated but only one corresponds to the image (HI), one is human-generated while the other is machine-generated (HM), both are machine-generated (MM). Each pair has 48 human judgments indicating which of the two candidates is more similar to a reference caption. Following previous work~\cite{hessel-etal-2021-clipscore} we take the majority vote for each candidate pair (ties are broken randomly) and randomly select 5 references out of the available 48 for reference-based metrics.

\begin{table}[t]
\small
\centering
\begin{small}
\begin{tabular}{cl}
\toprule
Cluster & Metrics \\
\midrule
& BLEU1, BLEU2, BLEU3, BLEU4, \\
1 & CIDEr, Exact NO, Fuzzy NO, \\
& METEOR, ROUGE, SPICE \\
\midrule
& BLIP2Score, CLIPImageScore, \\
2 & CLIPScore, MPNetScore, PACScore, \\
& RefCLIPScore, RefPACScore, Polos \\
\midrule
3 & Exact VO, Fuzzy VO \\
\bottomrule
    \end{tabular}
    \end{small}
    \caption{Metrics clustered by mutual correlation.}
    \label{tab:metric_clusters}
\end{table}

\datasetname{The Reformulations dataset}~\cite{berger2025improving} includes model-generated captions for 1,405 images from MSCOCO and Flickr30k, along with human-generated reformulations, i.e., corrected versions of the captions (if any errors exist). Evaluation metrics are expected to favor the caption after human correction. The dataset contains 5,208 caption-reformulation pairs; we omit 864 pairs where the reformulation is identical to the caption.

More details on the version we used for each dataset can be found in Appendix~\ref{sec:app_datasets}.

\myparagraph{Metrics.}
We select a subset of the 71 identified automatic metrics for evaluation. First, we exclude bias metrics, as bias is not captured by the human rating datasets. Next, we filter out all metrics that receive multiple candidates as input (such as diversity metrics), as the human ratings are provided for a single candidate. Finally, we exclude all metrics that rely on a closed model API access (CLAIR, GPT4V evaluation, ALOHa), focusing on publicly accessible evaluation methods.

After filtering, 56 metrics remain.
To ensure fair comparison, we conduct all the experiments for each metric ourselves, even if previous studies have reported results for this metric.
Due to the effort required and the scope of this study, we focus on metrics frequently employed in image captioning research. Specifically, we select metrics used in at least two papers per year on average since their publication.\footnote{Not including the publication year and 2024.} This narrows our selection to 20 metrics (Table~\ref{tab:metric_clusters}).\footnote{We omit two metrics for which we could not obtain a full implementation: PR-MCS and InfoMetIC.}
For details on the implementation of each metric, refer to Appendix~\ref{sec:app_metrics}.

\subsection{Metric Mutual Correlation} \label{sec:mutual_corr}

To examine relationships among the selected metrics, we analyze their mutual correlations using Spectral Clustering with $N=3$ clusters,\footnote{The number of clusters was chosen to maximize the Silhouette score.} where the adjacency between two metrics is defined as their mutual correlation on the Polaris train set. The resulting grouping is shown in Table~\ref{tab:metric_clusters}.
Cluster 1 primarily includes lexical similarity metrics, cluster 2 consists of Transformer-based metrics, and cluster 3 contains verb overlap metrics, which exhibit low correlation with all other metrics.

\subsection{\name{}: An Ensemble of Evaluation Methods}\label{subsec:ensemble}

\begin{table}[t]
\small
\centering
\begin{tabular}{lcc}
\toprule
Metric & Cluster & Coef \\
\midrule
Polos & 2 & 0.55 \\
BLIP2Score & 2 & 0.40 \\
PACScore & 2 & 0.29 \\
Exact NO & 1 & 0.08 \\
BLEU1 & 1 & 0.07 \\
Fuzzy VO & 3 & 0.02 \\
ROUGE & 1 & -0.07 \\
CIDEr & 1 & -0.15 \\
RefCLIPScore & 2 & -0.17 \\
\bottomrule
    \end{tabular}
    \caption{Metrics selected using feature selection to predict human ratings, their cluster numbers (as shown in Table~\ref{tab:metric_clusters}),
    and linear regression coefficients.}
    \label{tab:ensemble_weights}
\end{table}

\begin{table*}[t]
\small
\centering
\begin{tabular}{lccccc}
\toprule
Metric & Flickr8k & Flickr8k & Composite & THumB & Polaris \\
 & (Expert) & (CF) & & & \\
\midrule
BLEU4 & 30.8 & 16.9 & 30.6 & 10.4 & 46.2 \\
BLEU3 & 31.5 & 17.0 & 30.9 & 11.8 & 46.6 \\
BLEU2 & 32.5 & 17.9 & 31.1 & 15.8 & 47.1 \\
BLEU1 & 32.3 & 17.9 & 31.4 & 19.5 & 45.4 \\
ROUGE & 32.3 & 19.9 & 32.5 & 18.7 & 46.2 \\
METEOR & 41.8 & 22.3 & 39.0 & 18.5 & 51.2 \\
CIDEr & 43.9 & 24.6 & 37.5 & 22.4 & 52.1 \\
Dominant ensemble & 43.7 & 23.3 & 38.5 & 23.1 & 52.3 \\
SPICE & 44.9 & 24.4 & 40.4 & 21.0 & 50.9 \\
CLIPScore & 51.4 & 34.4 & 53.8 & 31.9 & 51.5 \\
PACScore & 54.3 & 36.0 & 55.7 & 31.4 & 52.4 \\
MPNetScore & 54.7 & 35.3 & 54.7 & 40.4 & 53.6 \\
RefCLIPScore & 53.0 & 36.4 & 55.4 & 40.7 & 54.1 \\
RefPACScore & 55.9 & 37.6 & 57.3 & 42.4 & 55.2 \\
BLIP2Score & 52.5 & 36.7 & 61.5 & 44.9 & 53.7 \\
Polos & 56.4 & 37.8 & 58.0 & 43.4 & 57.8 \\\midrule
\name{} (ours) & \textbf{58.5} & \textbf{38.7} & \textbf{61.7} & \textbf{46.6} & \textbf{58.8} \\
$\Delta$ from 2nd best & +2.1 & +0.9 & +0.2 & +1.7 & +1.0 \\
\bottomrule
    \end{tabular}
    \caption{Correlation with human ratings across different datasets. The best performing metric is in bold.
    }
    \label{tab:correlation_with_human}
\end{table*}

Our survey reveals that while a large number of metrics (71, as detailed in our literature review, Section 3) have been developed to assess diverse properties (outlined in our taxonomy, Section 4), only a small subset from a few categories is widely used (as shown in our usage analysis, Section 5). Furthermore, popular metric groups have faced increasing criticism in recent years. For instance, lexical similarity metrics are often criticized for failing to capture semantic similarity effectively ~\cite{gimenez-marquez-2007-linguistic}, while the recently popularized reference-free metrics have been challenged for focusing solely on visual grounding errors while neglecting caption implausibility errors ~\cite{ahmadi-agrawal-2024-examination}.

Consequently, we hypothesize that a more diverse evaluation approach, capable of capturing multiple aspects of a candidate caption, could enhance correlation with human ratings. As an initial step toward this goal, we explore a straightforward implementation: an ensemble method combining existing metrics through a linear combination.\footnote{Future research can explore more sophisticated multi-aspect metrics, which lies beyond the scope of this work.} We train the linear coefficients on one human rating dataset and evaluate the ensemble on additional datasets. If capturing multiple aspects indeed improves alignment with human judgments, the learned coefficients should generalize across datasets, resulting in improved correlation with human ratings. We refer to our proposed ensemble approach as \name{}.

Due to strong correlations among certain metrics, using all metrics in the ensemble might create redundancy. Therefore, we employ a sequential feature selection algorithm to identify a subset of metrics that accurately predict human ratings.
At each step, the algorithm adds the best metric to the ensemble based on the cross-validation score of a linear regression estimator of human ratings. The process stops when the estimator's score does not increase by at least $\varepsilon$. We use the Polaris train set\footnote{This is the only human rating dataset that is split into train/val/test.} for selecting the metrics and the validation set to determine the optimal $\varepsilon$ value (0.0001).

Next, we standardize the metrics values to the range $[0,1]$ and find the metrics coefficients by training a linear regression model to predict human ratings, again using the Polaris train dataset. The rescaling ensures that the coefficients are meaningfully comparable.\footnote{Rescaling did not impact the performance of our model.}
More details on our feature selection and linear regression procedures can be found in Appendix~\ref{sec:app_linear_regression}.

Table~\ref{tab:ensemble_weights} displays the selected metrics,
their cluster assignments (Section~\ref{sec:mutual_corr}),
and their linear regression coefficients.
The Transformer-based metrics (cluster 2) have the largest coefficients, making them the most influential in predicting the candidate score. Negative coefficients are observed for some metrics in clusters 1 and 2, likely to mitigate redundancy from multiple metrics within the same cluster. For instance, BLIP2Score, Polos, and PACScore all receive positive coefficients but are highly correlated. To balance this redundancy, a fourth metric from the same cluster (RefCLIPScore) is assigned a negative coefficient.

Considering the frequent use of the five dominant metrics in prior research, we include an ensemble of these metrics (the dominant ensemble) as a baseline. As before, we determine the coefficients for this ensemble using linear regression.

\begin{table*}[t]
\small
\centering
\begin{tabular}{ccccccc}
\toprule
\multirow{2}{*}{Metric} & \multicolumn{5}{c}{Pascal50S} & \multirow{2}{*}{Reformulations} \\
& HC & HI & HM & MM & Mean & \\
\midrule
BLEU4 & 59.8 $\pm$ 0.7 & 92.3 $\pm$ 0.7 & 85.6 $\pm$ 0.5 & 57.3 $\pm$ 1.2 & 73.7 $\pm$ 0.5 & 49.8 \\
BLEU3 & 60.6 $\pm$ 0.9 & 93.0 $\pm$ 0.7 & 88.1 $\pm$ 0.5 & 57.6 $\pm$ 1.0 & 74.8 $\pm$ 0.6 & 51.8 \\
ROUGE & 63.1 $\pm$ 0.8 & 95.2 $\pm$ 0.6 & 91.9 $\pm$ 0.3 & 60.3 $\pm$ 1.0 & 77.6 $\pm$ 0.4 & 53.1 \\
BLEU2 & 62.9 $\pm$ 1.7 & 94.0 $\pm$ 0.4 & 90.1 $\pm$ 0.2 & 58.6 $\pm$ 1.0 & 76.4 $\pm$ 0.7 & 54.9 \\
BLEU1 & 62.8 $\pm$ 1.3 & 95.1 $\pm$ 0.5 & 91.5 $\pm$ 0.4 & 59.8 $\pm$ 0.7 & 77.3 $\pm$ 0.2 & 56.0 \\
CIDEr & 65.6 $\pm$ 1.7 & 98.0 $\pm$ 0.4 & 90.9 $\pm$ 0.3 & 64.9 $\pm$ 1.1 & 79.8 $\pm$ 0.4 & 54.3 \\
SPICE & 61.2 $\pm$ 1.9 & 94.3 $\pm$ 0.7 & 85.5 $\pm$ 0.7 & 49.5 $\pm$ 0.7 & 72.6 $\pm$ 0.7 & 67.6 \\
METEOR & 64.4 $\pm$ 1.7 & 97.6 $\pm$ 0.4 & 94.1 $\pm$ 0.8 & 65.3 $\pm$ 0.9 & 80.4 $\pm$ 0.6 & 66.3 \\
Dominant ensemble & 65.5 $\pm$ 1.7 & 97.7 $\pm$ 0.3 & 93.0 $\pm$ 0.5 & 68.0 $\pm$ 0.7 & 81.1 $\pm$ 0.6 & 66.0 \\
RefPACScore & 67.6 $\pm$ 0.7 & 99.6 $\pm$ 0.1 & 96.0 $\pm$ 0.2 & 75.7 $\pm$ 0.6 & 84.7 $\pm$ 0.3 & 73.3 \\
RefCLIPScore & 64.1 $\pm$ 1.2 & 99.6 $\pm$ 0.1 & 95.8 $\pm$ 0.4 & 72.9 $\pm$ 0.6 & 83.1 $\pm$ 0.4 & 74.9 \\
BLIP2Score & 60.5 $\pm$ 0.2 & \textbf{99.8 $\pm$ 0.0} & 96.3 $\pm$ 0.0 & 71.2 $\pm$ 0.5 & 82.0 $\pm$ 0.1 & 76.1 \\
Polos & 69.5 $\pm$ 1.2 & 99.6 $\pm$ 0.0 & 97.0 $\pm$ 0.3 & 78.5 $\pm$ 0.6 & 86.1 $\pm$ 0.1 & 73.0 \\
MPNetScore & \textbf{71.9 $\pm$ 0.6} & 99.8 $\pm$ 0.1 & 96.3 $\pm$ 0.5 & \textbf{79.0 $\pm$ 0.6} & \textbf{86.7 $\pm$ 0.2} & 72.7 \\
PACScore & 60.4 $\pm$ 0.1 & 99.3 $\pm$ 0.0 & 96.8 $\pm$ 0.0 & 72.9 $\pm$ 0.4 & 82.4 $\pm$ 0.1 & 77.2 \\
CLIPScore & 56.1 $\pm$ 0.2 & 99.3 $\pm$ 0.0 & 96.3 $\pm$ 0.0 & 71.2 $\pm$ 0.4 & 80.7 $\pm$ 0.1 & 80.8 \\
\name{} (ours) & 68.7 $\pm$ 0.6 & 99.8 $\pm$ 0.1 & \textbf{98.3 $\pm$ 0.3} & 77.3 $\pm$ 0.3 & 86.0 $\pm$ 0.2 & \textbf{81.4} \\
\bottomrule
    \end{tabular}
    \caption{Accuracy in pairwise comparison. In Pascal50S we report mean and standard deviation across five random instances of tie-breaking and reference selection. HC: both captions are human-generated for the image, HI: both are human-generated but only one corresponds to the image, HM: one caption is human-generated while the other is machine-generated, MM: both captions are machine-generated. In each dataset, best scoring metric is in bold.
    }
    \label{tab:pairwise_comparison}
\end{table*}

\subsection{Correlation with Human Ratings} \label{sec:corr_with_human}

We now compute the correlation between metric scores and human ratings across all datasets containing human ratings (all but Pascal50S and the Reformulations datasets).\footnote{For Polaris we use the test set.} We follow previous work~\cite{anderson2016spice, hessel-etal-2021-clipscore, kasai-etal-2022-transparent, Wada_2024_CVPR} and use Kendall-C correlation for Flickr8k-Expert, Composite and Polaris, Kendall-B correlation for Flickr8k-CF and Pearson correlation for THumB.

Table~\ref{tab:correlation_with_human} presents the results. We omit results for five metrics\footnote{CLIPImageScore, Exact NO, Exact VO, Fuzzy NO, Fuzzy VO.} that were proposed ad hoc in an experiment (rather than in a dedicated paper) and showed weak performance.

\myparagraph{An ensemble improves over individual metrics.}
\name{} demonstrates the highest correlation with human ratings across all datasets, indicating that integrating various aspects of caption quality yields a more ``human-like'' score.

\myparagraph{Dominant metrics are weaker than recent alternatives.}
The five dominant metrics, as well as their ensemble, correlate less with human ratings compared to all other examined metrics.

\myparagraph{Lesser-known metrics prove to be valuable.}
While some metrics were introduced in dedicated papers, others were proposed ad hoc in the experiments section. Previous studies comparing metrics performance have only discussed the former, missing strong correlations with human ratings observed with ad hoc metrics like MPNetScore and BLIP2Score. This underscores the importance of a systematic review of all relevant papers.

\subsection{Accuracy on Pairwise Comparison}

We also perform a pairwise comparison task on the Pascal50S and Reformulations datasets to assess metrics' accuracy in assigning a higher score to the human-preferred candidate in each pair. For Pascal50S, we use majority vote to indicate human preference and report the mean and standard deviation across five random instances of tie-breaking and reference selection. For Reformulations, we consider the reformulated caption as preferred.

Results are presented in Table~\ref{tab:pairwise_comparison}. \name{} attained the highest score on the Reformulations dataset, as well as when comparing human- and machine-generated captions in Pascal 50 (HM). In all other cases MPNetScore and BLIP2Score performed best. As noted earlier, these metrics were absent from previous studies since they were not introduced in dedicated papers.

\subsection{Why Does the Ensemble Improve Correlation?}

To understand why \name{} improves correlation over individual metrics, we manually examine cases where scores of individual metrics deviate from human ratings while the ensemble score aligns more closely.

\begin{figure}[tb]
    \centering
    \includegraphics[width=\columnwidth]{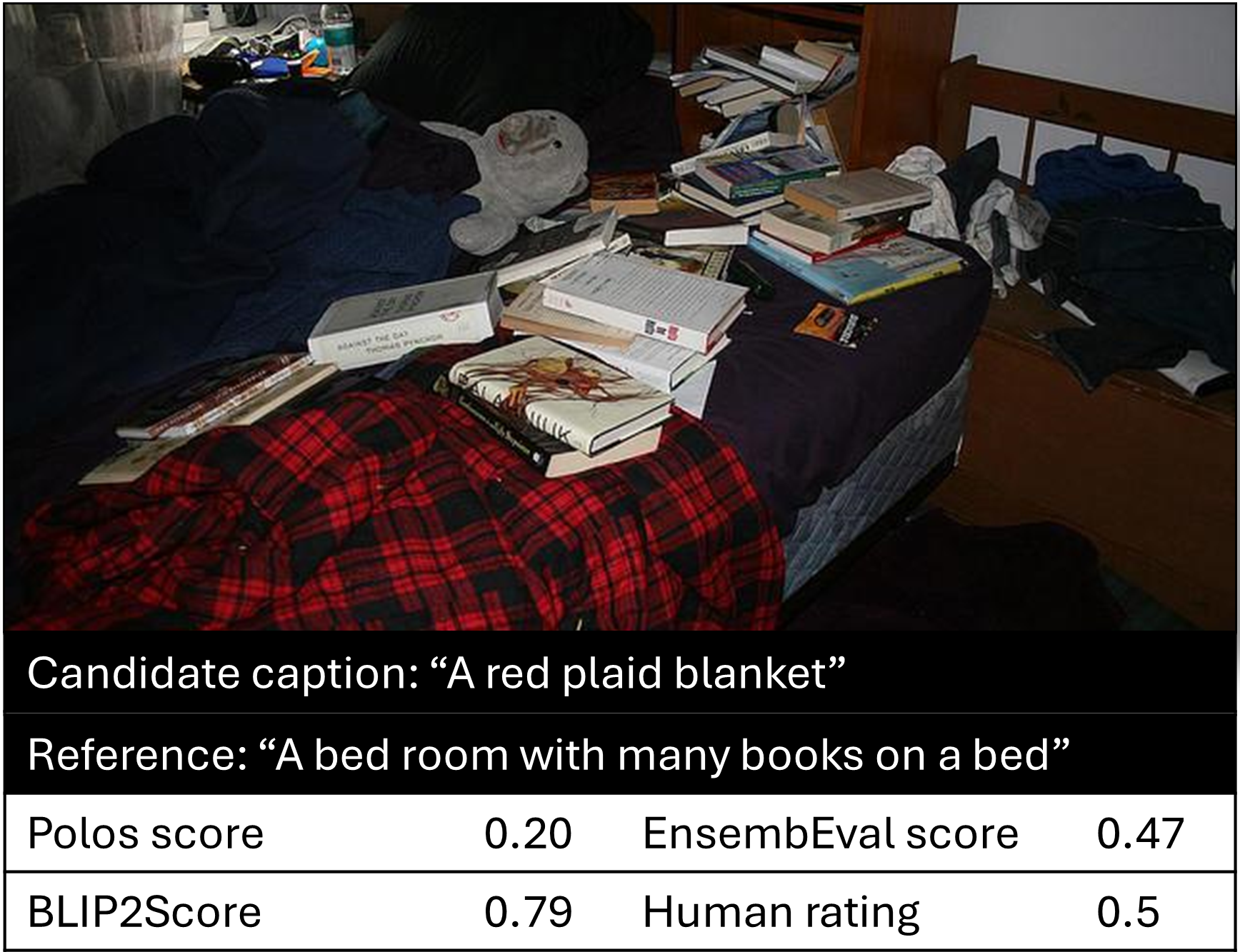}
    \caption{An image, corresponding candidate caption, reference example and evaluation scores. Since no reference captions mention the blanket, Polos (a reference-based metric) assigns a low score. However, as the caption accurately describes the image, BLIP2Score (a reference-free metric) gives a high score.
    The ensemble score falls in between, aligning best with human ratings. Image and caption taken from the Polaris dataset.}
    \label{fig:ensemble_example}
\end{figure}

These cases often arise from disagreements on which entities to describe. If reference captions omit an entity mentioned in the candidate caption, reference-based metrics assign a low score, while reference-free metrics give a high one. Humans typically rate such captions moderately, as they are accurate but may not align with their descriptive choices. The ensemble, combining both metric types, aligns more closely with human ratings. This issue is especially relevant beyond English, as research shows speakers of different languages tend to describe different entities~\cite{berger2024cross}.
See Figure~\ref{fig:ensemble_example} for an example.

\section{Software Package} \label{sec:sw_package}

\newmaterial{Our empirical analysis of the redundancy/complementarity of the different metrics, reveals that an ensemble of several methods may be considerably better correlated with human rankings of the captions.
To this aim, we release a software package \name{} under the permissive MIT license.\footnote{\iftaclpubformat
\href{https://github.com/uriberger/caption_evaluation}{github.com/uriberger/caption\_evaluation}
\else
\href{https://anonymous.4open.science/r/caption_evaluation-6021/}{anonymous.4open.science/r/caption\_evaluation-6021}
\fi} 

The software package implements the ensemble metric presented in Section \ref{subsec:ensemble}. It also straightforwardly supports the use of alternative sets of weights. The software package can easily be extended with additional metrics, either reference-based or reference-free. The package is therefore usable not only for replication of the results presented in this paper, but also as a generic tool for ensembling image captioning evaluation metrics.
}

\section{Related Work} \label{sec:related_work}
\subsection{Image Captioning Surveys}

To the best of our knowledge, no previous work has exclusively surveyed image captioning evaluation methods. However, most image captioning surveys include a dedicated section for automatic evaluation methods, typically mentioning the five dominant metrics along with one or two additional ones~\citep[e.g.,][]{hossain2019comprehensive, liu2019survey, ghandi2023deep, sharma2023comprehensive}.

A few studies~\cite{stefanini2022show, XU2023126287} provide a more detailed taxonomy of automatic metrics. Our work differs from them in two ways: First, our systematic review identifies strong metrics not covered previously.
Second, while these studies group metrics into loosely defined categories (e.g., ``standard'' metrics), we clearly define our categorization criteria based on the properties each metric aims to measure.

We are also the first to develop a taxonomy for human evaluation frameworks. Before our study, \citet{bernardi2016automatic} discussed human evaluation but only mentioned the scale rating framework.

Prior to our study, \citet{staniute2019systematic} reported metric usage from selected papers; however, they manually selected popular papers rather than identifying them systematically.
Similar to our work, two studies~\cite{sharma2021survey, Al-Shamayleh2024} systematically identify image captioning papers and report metric usage, but cover much fewer papers (79 and 41, respectively) and metrics (3 and 8, respectively), and do not report trends or provide in-depth analysis.

\subsection{Image Captioning Metrics Analysis}

Several previous studies have analyzed the effectiveness of image captioning metrics.
\citet{hodosh2013framing} compare BLEU and ROUGE scores to human ratings and find that these metrics fall short in measuring content quality.
\citet{elliott-keller-2014-comparing} compute the correlation of lexical similarity metrics with human ratings and find that METEOR had the strongest correlation.
\citet{kilickaya-etal-2017-evaluating} compare the five dominant metrics and WMD, finding that $n$-gram based metrics exhibit lower performance than SPICE and WMD.

Recently, as reference-free metrics like CLIPScore gain prominence, there has been growing criticism directed at this line of research.
\citet{deutsch-etal-2022-limitations} argue that reference-free metrics can be exploited at test time to find outputs that maximize their scores.
\citet{ahmadi-agrawal-2024-examination} find these metrics sensitive to visual grounding errors but not to caption implausibility.

\section{Conclusion}
We conducted a comprehensive survey of image captioning evaluation methods, yielding two key outcomes: a taxonomy of captioning metrics, including overlooked high-performing ones (e.g., BLIP2Score) and an analysis revealing that most papers rely on five metrics that have only a weak correlation with human ratings. We further showed that an ensemble of several existing metrics improves correlation with human ratings, and proposed a simple weighted ensemble method to this effect. We release a software package that implements this ensemble methods and facilitates future developments in image captioning metric ensembling methods.



\iftaclpubformat
\section*{Acknowledgments}
This work was supported in part by the Israel Science Foundation (grant no. 2424/21), by a grant from the Israeli Planning and Budgeting Committee (PBD) and by the HUJI-UoM joint PhD program.
\fi

\bibliography{tacl2021, custom, anthology}
\bibliographystyle{acl_natbib}

\onecolumn

\appendix
\section{Experiments: Additional Information} \label{sec:app_experiments}
\subsection{Datasets} \label{sec:app_datasets}

\paragraph{Flickr8k.}
We use the official version of the Flickr8k dataset.\footnote{\href{https://www.kaggle.com/datasets/dibyansudiptiman/flickr-8k}{www.kaggle.com/datasets/dibyansudiptiman/flickr-8k}} We follow previous work~\cite{anderson2016spice, hessel-etal-2021-clipscore} when handling references that are also included in the candidate set: in Flickr8k-Expert we remove these sentences from the candidate set (158 examples), and in Flickr8k-CF we remove these sentences from the reference set.

In Flickr8k-expert, annotators are instructed to rank the candidate on a scale of 1 to 4, with the following specifications:

4 = Sentence describes the image: Sentence contains no errors, everything described in the sentence appears in the image.

3 = Sentence almost describes the image: Sentence contains only a single moderate error or a small number of minor errors, major details described in the sentence appear in the image.

2 = Sentence barely describes the image: Sentence contains a major error or many moderate or minor errors, only some minor details described in the sentence appear in the image.

1 = Sentence does not describe the image at all: No details described in the sentence appear in the image, sentence is unrelated to image.

In Flickr8k-CF, annotators are shown images with ten sentences below each images, and are asked to select all sentences that could describe the image, with the following specifications:

Yes- The sentence is a good enough description of the image (minor details may be wrong, and not everything that appears in the image has to be described).

No- The sentence does not describe the image because it contains major errors or is completely unrelated to the image.

\paragraph{Composite.}
We use the official version of the Composite dataset.\footnote{\href{https://imagesdg.wordpress.com/image-to-scene-description-graph/}{imagesdg.wordpress.com/image-to-scene-description-graph}}
While previous work mentioned that there are three candidates per image and a single score per candidate, we find that the dataset contains four candidates per image for MSCOCO and Flickr30k, and two scores per candidate (correctness and throughness). We find that results from prior work was most closely replicated when we discarded the fourth candidate when one exists and used the correctness score as the total score.
Also, following \citet{hessel-etal-2021-clipscore} we remove sentences that appear both as references and as candidates from the reference set.

Instructions for the correctness rating (which we use in this study) are as follows:

Descriptions that correctly describe the image content with higher precision have better correctness ratings.

Score 1: The description has no relevance to the image.

Score 2: The description have only weak relevance to the image.

Score 3: The description have some relevance to the image.

Score 4: The description relates closely to the image.

Score 5: The description relates perfectly to the image.

\paragraph{THumB.}
We use the official version of the THumB dataset.\footnote{\href{https://github.com/jungokasai/THumB}{github.com/jungokasai/THumB}}
Following the practice in other datasets, we remove sentences that appear both as references and as candidates from the reference set.

In this study, the authors performed the annotations themselves, relying on general descriptions for each aspect of the rating rather than providing specific instructions.

\paragraph{Polaris.}
We use the HuggingFace version of the Polaris dataset.\footnote{\href{https://huggingface.co/datasets/yuwd/Polaris}{huggingface.co/datasets/yuwd/Polaris}}

Annotators were instructed to rate the candidate on a scale of 1 to 5, with the following specifications:

Evaluate the extent to which the sentences capture the features of the image, considering fluency, relevance and descriptiveness.

Use a five-point scale for your evaluation, employing a point addition or deduction system.

For descriptiveness, rate a caption as `Excellent' if it is both detailed and accurate.

For relevance, award points if the caption accurately depicts any present object, even if irrelevant words are included. Deduct points for each irrelevant word.

For fluency, assign lower scores if there are grammatical errors.

\begin{wrapfigure}{r}{0.4\textwidth}
    \centering
    \includegraphics[width=5cm]{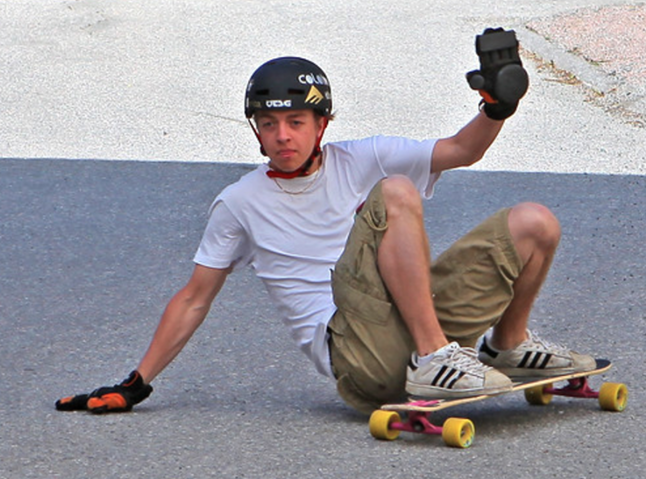}
    \caption{Image provided as an example in the Polaris dataset instructions.}
    \vspace{-4ex}
    \label{fig:polos_image}
\end{wrapfigure}

For example, consider the caption `A man wearing a helmet is skateboarding with a dog behind him.' with the given image (Figure~\ref{fig:polos_image}). This caption scores high in relevance as it accurately describes the main object in the image. However, the mention of `a dog,' which is absent from the image, reduces its descriptiveness. Despite its high fluency, this warrants a 'Fair' (3 on a scale of 1 to 5) rating overall.

\paragraph{Pascal50S.}
The link to download the official version of the dataset in the CIDEr blog post {\href{https://vrama91.github.io/cider/}{[Link]}} seems to be broken; we thank
(REMOVED FOR ANONIMYZATION)
for providing us with the original files.

\subsection{Metrics implementations.} \label{sec:app_metrics}

\paragraph{The five dominant metrics.}
For the five most dominant metrics we use the pycocoevalcap implementation. {\href{https://pypi.org/project/pycocoevalcap/}{[Link]}

\paragraph{Exact/fuzzy noun/verb overlap.}
We use an implementation {\href{https://github.com/DavidMChan/caption-by-committee/blob/main/cbc/metrics/content_score.py}{[Link]}  provided to us by the author.

\paragraph{CLIPImageScore.}
Since no implementation is provided by the authors, we implement the metric ourselves. We use the SD-XL 1.0-base model {\href{https://huggingface.co/stabilityai/stable-diffusion-xl-base-1.0}{[Link]]}} to generate the image based on the candidate.

\paragraph{(Ref) CLIP and PAC Score, Polos.}
We use the official implementations. {\href{https://github.com/jmhessel/clipscore}{[CLIP]}} {\href{https://github.com/aimagelab/pacscore}{[PAC]}} {\href{https://github.com/keio-smilab24/Polos}{[Polos]}}

\paragraph{BLIP2Score.}
Since no implementation is provided by the authors, we implement the metric ourselves. We use the BLIP2 image-text-matching model from the LAVIS package~\cite{li-etal-2023-lavis}.

\paragraph{MPNetScore.}
Since no implementation is provided by the authors, we implement the metric ourselves. We use the all-mpnet-base-v2 model. {\href{https://huggingface.co/sentence-transformers/all-mpnet-base-v2}{[Link]}}

\subsection{Feature Selection and Linear Regression} \label{sec:app_linear_regression}
We use the scikit-learn {\href{https://scikit-learn.org/stable/}{[Link]}} implementations of feature selection and linear regression. We use forward-selection in feature selection.

\subsection{Reproducibility Issues} \label{sec:app_reproducibility}
In our experiments in Section~\ref{sec:corr_with_human} we tried our best to replicate the results reported by previous studies. We now describe the cases where we were enable to. We compare our results to \citet{hessel-etal-2021-clipscore}, \citet{kasai-etal-2022-transparent}, \citet{Sarto_2023_CVPR}, \citet{hu-etal-2023-infometic} and \citet{Wada_2024_CVPR}.

\paragraph{The five dominant metrics.}
Across some of the datasets, we found small differences between our values and previous reported values, no larger than 0.2. One exception is the THumB dataset, where prior work~\cite{kasai-etal-2022-transparent, hu-etal-2023-infometic} report no digits after the decimal point, so the differences are no larger than 1.0, except for BLEU4, where the difference was larger because the authors used the sacrebleu implemenation while we used the pycocoevalcap implementation, consistent with the common practice in image captioning.

\paragraph{(Ref) CLIP and PAC Score.}
Across all datasets our results are identical to those previous reported.
%
One exception is the Polaris dataset, where our results for these metrics differed from those of \citet{Wada_2024_CVPR}. Since we used the HuggingFace version of the dataset and the official implementations for these metrics, we assume this discrepancy results from the authors using a more preliminary version of the dataset.

\newpage

\section{Metric Usage Records} \label{sec:app_metric_usage}
We now list all the papers we examined and the metric categories they used. Abbreviations: LexSim: Lexical similarity, RefPhSim: Candidate-reference phrase semantic similarity, RefSenSim: Candidate-reference sentence-level similarity, InfoSim: Extracted information similarity, ImPhSim: Candidate-image phrase-level semantic similarity, SenImSim: Sentence-image similarity, Ret: Retrieval-based methods, MulSim: Candidate-multiple source similarity, Flu: Fluency, Div: Diversity, Yngve: Yngve score, Bias: Bias.

\paragraph{2024.}
\emph{NLP} \citet{yang-etal-2024-masking}: LexSim, RefPhSim, Bias, \citet{ramos-etal-2024-paella}: LexSim, \citet{tanaka-etal-2024-content}: Div, \citet{qu-etal-2024-visually}: LexSim, \citet{li-etal-2024-role}: LexSim, SenImSim

\emph{Vision} \citet{Ge_2024_CVPR}: RefSenSim, SenImSim, \citet{Huang_2024_CVPR}: LexSim, RefPhSim, \citet{zeng2024meacap}: LexSim, RefPhSim, SenImSim, \citet{Li_2024_CVPR}: LexSim, RefPhSim

\emph{Machine Learning} \citet{ma2024sycoca}: LexSim, RefPhSim, \citet{black2024vixen}: LexSim, RefSenSim, \citet{Liu_Liu_Ma_2024}: LexSim, RefPhSim, \citet{Ma_Zhou_Rao_Zhang_Sun_2024}: LexSim, \citet{Qi_Zhao_Wu_2024}: LexSim, \citet{Qiu_Ning_He_2024}: LexSim, RefPhSim, \citet{Wang_Deng_Jia_2024}: LexSim, RefPhSim, \citet{Fu_Song_Zhou_Yang_2024}: LexSim, RefPhSim

\paragraph{2023.}
\emph{NLP} \citet{wang-etal-2023-improving}: LexSim, RefPhSim, \citet{qiu-etal-2023-gender}: LexSim, RefPhSim, SenImSim, \citet{hwang-shwartz-2023-memecap}: LexSim, RefPhSim, \citet{chan-etal-2023-ic3}: LexSim, RefPhSim, Ret, \citet{maeda-etal-2023-query}: LexSim, RefPhSim, SenImSim, MulSim, \citet{yang-etal-2023-vilm}: LexSim, RefPhSim, \citet{mohamed-etal-2023-violet}: LexSim, \citet{wu-etal-2023-cross2stra}: LexSim, \citet{yang-etal-2023-multicapclip}: LexSim, RefPhSim, \citet{yang-etal-2023-transforming}: LexSim, RefPhSim, \citet{rajakumar-kalarani-etal-2023-lets}: LexSim, \citet{ramos-etal-2023-lmcap}: LexSim, \citet{ou-etal-2023-pragmatic}: Flu, \citet{zhang-wan-2023-exploring}: LexSim, \citet{bielawski-vanrullen-2023-clip}: LexSim, \citet{anagnostopoulou-etal-2023-towards}: LexSim, \citet{zhao-etal-2023-generating-visual}: LexSim, RefPhSim, \citet{ramos-etal-2023-retrieval}: LexSim, RefPhSim, \citet{zhou-long-2023-style}: LexSim, RefPhSim, Flu

\emph{Vision} \citet{Wu_2023_ICCV}: LexSim, \citet{Li_2023_ICCV}: Flu, Div, \citet{Kornblith_2023_ICCV}: LexSim, SenImSim, Ret, MulSim, \citet{Hu_2023_ICCV}: LexSim, RefPhSim, \citet{Tu_2023_ICCV}: LexSim, RefPhSim, \citet{Fei_2023_ICCV}: LexSim, RefPhSim, \citet{Fan_2023_ICCV}: LexSim, RefPhSim, \citet{Kang_2023_ICCV}: LexSim, RefPhSim, SenImSim, Ret, \citet{Barraco_2023_ICCV}: LexSim, RefPhSim, SenImSim, MulSim, \citet{Hu_2023_ICCV}: LexSim, RefPhSim, \citet{Chen_2023_ICCV}: LexSim, \citet{Dessi_2023_CVPR}: LexSim, RefPhSim, \citet{Vo_ACAP_2023_CVPR}: LexSim, RefPhSim, SenImSim, Ret, MulSim, \citet{Luo_2023_CVPR}: LexSim, RefPhSim, \citet{Zeng_2023_CVPR}: LexSim, RefPhSim, SenImSim, MulSim, Div, \citet{Kuo_2023_CVPR}: LexSim, RefPhSim, \citet{Chen_2023_CVPR}: LexSim, \citet{Ramos_2023_CVPR}: LexSim, RefPhSim, \citet{Hirota_2023_CVPR}: LexSim, RefPhSim, SenImSim, Bias, \citet{Ren_2023_CVPR}: LexSim, RefPhSim

\emph{Machine Learning} \citet{NEURIPS2023_804b5e30}: LexSim, \citet{NEURIPS2023_fa1cfe4e}: SenImSim, Ret, Flu, Div, \citet{NEURIPS2023_ee4814f9}: SenImSim, Ret, \citet{pmlr-v202-zheng23g}: LexSim, \citet{li2023decapdecodingcliplatents}: LexSim, RefPhSim, \citet{Fei_Fan_Zhu_Huang_Wei_Wei_2023}: LexSim, RefPhSim, \citet{Zhong_Zhou_Qiu_2023}: LexSim, RefPhSim, \citet{Nguyen_Biten_Mafla_Gomez_Karatzas_2023}: LexSim, RefPhSim, \citet{Wang_Xie_Luo_Cheng_Wu_Jia_Li_2023}: LexSim, RefPhSim, \citet{Wang_Xie_Wu_Jia_Li_2023}: LexSim, RefPhSim

\paragraph{2022.}
\emph{NLP} \citet{gao-etal-2022-caponimage}: LexSim, Div, \citet{zhang-etal-2022-cross}: LexSim, RefPhSim, \citet{mirchandani-etal-2022-fad}: LexSim, \citet{nukrai-etal-2022-text}: LexSim, \citet{zhou-etal-2022-focus}: LexSim, \citet{zhao-etal-2022-visual}: LexSim, RefPhSim, \citet{nishino-etal-2022-factual}: LexSim, RefPhSim, InfoSim, \citet{cafagna-etal-2022-understanding}: LexSim, RefPhSim, \citet{cho-etal-2022-fine}: LexSim, RefPhSim, SenImSim, Ret, MulSim, \citet{pantazopoulos-etal-2022-combine}: LexSim, RefPhSim, \citet{xu-etal-2022-joint}: LexSim, \citet{guo-etal-2022-clip4idc}: LexSim

\emph{Vision} \citet{10.1007/978-3-031-19833-5_31}: LexSim, \citet{10.1007/978-3-031-19833-5_41}: LexSim, RefPhSim, \citet{10.1007/978-3-031-20074-8_13}: LexSim, \citet{10.1007/978-3-031-20059-5_7}: LexSim, RefPhSim, \citet{10.1007/978-3-031-20059-5_10}: LexSim, RefPhSim, \citet{10.1007/978-3-031-20059-5_13}: LexSim, RefPhSim, \citet{Chen_2022_CVPR}: LexSim, \citet{Hirota_2022_CVPR}: LexSim, Bias, \citet{Cai_2022_CVPR}: LexSim, \citet{Wu_2022_CVPR}: LexSim, RefPhSim, \citet{Chen_2022_CVPR}: LexSim, \citet{Yuan_2022_CVPR}: LexSim, \citet{Li_2022_CVPR}: LexSim, RefPhSim, \citet{Fei_2022_CVPR}: LexSim, RefPhSim, \citet{Liu_2022_CVPR}: LexSim, RefPhSim, \citet{Vo_2022_CVPR}: LexSim, RefPhSim, \citet{Mohamed_2022_CVPR}: LexSim, \citet{Fang_2022_CVPR}: LexSim, RefPhSim, \citet{Hu_2022_CVPR}: LexSim, RefPhSim, \citet{Kuo_2022_CVPR}: LexSim, RefPhSim, \citet{Mavroudi_2022_CVPR}: LexSim, RefPhSim

\emph{Machine Learning} \citet{NEURIPS2022_2a8e6c09}: LexSim, RefPhSim, \citet{NEURIPS2022_2a8e6c09}: LexSim, RefPhSim, SenImSim, Div, \citet{NEURIPS2022_6e4df340}: ImPhSim, \citet{Fei_2022}: LexSim, RefPhSim, \citet{Zhang_Shi_Guo_Zhang_Cai_Li_Luo_Zhuang_2022}: LexSim, RefPhSim, Div, \citet{Yao_Wang_Jin_2022}: LexSim, \citet{Wang_Xu_Sun_2022}: LexSim, RefPhSim, \citet{Feng_Lu_Tao_Alikhani_Mitamura_Hovy_Gangal_2022}: LexSim, RefPhSim, Flu

\paragraph{2021.}
\emph{NLP} \citet{yang-etal-2021-journalistic}: LexSim, \citet{liu-etal-2021-visual}: LexSim, \citet{tu-etal-2021-r}: LexSim, RefPhSim, \citet{chen-etal-2021-language-resource}: LexSim, \citet{shi-etal-2021-retrieval-analogy}: LexSim, RefPhSim, Div, \citet{ng-etal-2021-understanding}: LexSim, RefPhSim, Div, \citet{yan-etal-2021-control}: LexSim, \citet{shi-etal-2021-enhancing}: LexSim, RefPhSim, Ret, \citet{tu-etal-2021-semantic}: LexSim, RefPhSim, \citet{bugliarello-elliott-2021-role}: LexSim, RefPhSim, \citet{wang-etal-2021-ecol}: LexSim, RefPhSim, \citet{honda-etal-2021-removing}: LexSim, RefPhSim, \citet{nag-chowdhury-etal-2021-exploiting}: LexSim, RefPhSim, \citet{ahsan-etal-2021-multi}: LexSim, RefPhSim, \citet{zhong-miyao-2021-leveraging}: LexSim, RefPhSim, \citet{nagasawa-etal-2021-validity}: LexSim, RefPhSim

\emph{Vision} \citet{Shi_2021_ICCV}: LexSim, Div, \citet{Kim_2021_ICCV}: LexSim, RefPhSim, \citet{Zhao_2021_ICCV}: LexSim, RefPhSim, \citet{Yang_2021_ICCV}: LexSim, RefPhSim, \citet{Nguyen_2021_ICCV}: LexSim, RefPhSim, \citet{Yang_2021_CVPR}: LexSim, \citet{Chen_2021_CVPR}: LexSim, RefPhSim, Div, \citet{Hosseinzadeh_2021_CVPR}: LexSim, RefPhSim, \citet{Xu_2021_CVPR}: LexSim, RefPhSim, Div, \citet{Wang_2021_CVPR_b}: LexSim, RefPhSim, \citet{Chen_2021_CVPR_b}: LexSim, \citet{Zhang_2021_CVPR}: LexSim, RefPhSim

\emph{Machine Learning} \citet{Hu_Yin_Lin_Zhang_Gao_Wang_Liu_2021}: LexSim, RefPhSim, \citet{Ji_Luo_Sun_Chen_Luo_Wu_Gao_Ji_2021}: LexSim, RefPhSim, \citet{Fei_2021}: LexSim, RefPhSim, \citet{Fei_2021_b}: LexSim, RefPhSim, \citet{Song_Zhou_Mao_Tan_2021}: LexSim, RefPhSim, \citet{Luo_Ji_Sun_Cao_Wu_Huang_Lin_Ji_2021}: LexSim, RefPhSim, \citet{Zhang_Shi_Tang_Xiao_Yu_Zhuang_2021}: LexSim, RefPhSim, \citet{Yang_Yang_Hsu_2021}: LexSim, \citet{Kim_Zala_Burri_Bansal_2021}: LexSim

\paragraph{2020.}
\emph{NLP} \citet{milewski-etal-2020-scene}: LexSim, RefPhSim, \citet{alikhani-etal-2020-cross}: LexSim, \citet{shi-etal-2020-improving}: LexSim, RefPhSim, \citet{park-etal-2020-feature}: LexSim, \citet{nie-etal-2020-pragmatic}: LexSim, \citet{nguyen-etal-2020-structural}: LexSim, RefPhSim, \citet{takmaz-etal-2020-generating}: LexSim, RefPhSim

\emph{Vision} \citet{10.1007/978-3-030-58452-8_22}: LexSim, RefPhSim, Ret, Div, \citet{10.1007/978-3-030-58536-5_44}: LexSim, RefPhSim, \citet{10.1007/978-3-030-58571-6_37}: LexSim, RefPhSim, \citet{10.1007/978-3-030-58601-0_1}: LexSim, RefPhSim, \citet{10.1007/978-3-030-58601-0_42}: LexSim, RefPhSim, \citet{10.1007/978-3-030-58568-6_13}: LexSim, RefPhSim, Div, \citet{10.1007/978-3-030-58568-6_34}: LexSim, RefPhSim, \citet{10.1007/978-3-030-58520-4_25}: LexSim, RefPhSim, \citet{10.1007/978-3-030-58523-5_21}: LexSim, RefPhSim, \citet{Chen_2020_CVPR}: LexSim, RefPhSim, \citet{Sammani_2020_CVPR}: LexSim, RefPhSim, \citet{Guo_2020_CVPR}: LexSim, RefPhSim, \citet{Chen_2020_CVPR_b}: LexSim, RefPhSim, Div, \citet{Cornia_2020_CVPR}: LexSim, RefPhSim, \citet{Pan_2020_CVPR}: LexSim, RefPhSim, \citet{Zhou_2020_CVPR}: LexSim, RefPhSim, \citet{Tran_2020_CVPR}: LexSim

\emph{Machine Learning} \citet{NEURIPS2020_c2964caa}: LexSim, \citet{NEURIPS2020_24bea84d}: LexSim, RefPhSim, Div, \citet{Zhang_Ying_Lu_Zha_2020}: LexSim, RefPhSim, \citet{Zhou_Palangi_Zhang_Hu_Corso_Gao_2020}: LexSim, RefPhSim, \citet{Zhao_Wu_Zhang_2020}: LexSim, Flu, \citet{Wang_Bai_Zhang_Lu_2020}: LexSim, RefPhSim, \citet{Liu_Wang_Xu_Zhao_Xu_Shen_Yang_2020}: LexSim, \citet{Hou_Wu_Zhang_Qi_Jia_Luo_2020}: LexSim, RefPhSim, \citet{Cao_Han_Wang_Ma_Fu_Jiang_Xue_2020}: LexSim, RefPhSim

\paragraph{2019.}
\emph{NLP} \citet{zhao-etal-2019-informative}: LexSim, \citet{fan-etal-2019-bridging}: LexSim, \citet{nikolaus-etal-2019-compositional}: LexSim, RefPhSim, \citet{changpinyo-etal-2019-decoupled}: LexSim, RefPhSim, \citet{kim-etal-2019-image}: LexSim, RefPhSim, \citet{wang-etal-2019-role}: LexSim, RefPhSim

\emph{Vision} \citet{Shen_2019_ICCV}: LexSim, Div, \citet{Liu_2019_ICCV}: LexSim, RefPhSim, Ret, Div, Yngve, \citet{Ge_2019_ICCV}: LexSim, \citet{Yao_2019_ICCV}: LexSim, RefPhSim, \citet{Yang_2019_ICCV}: LexSim, RefPhSim, \citet{Aneja_2019_ICCV}: LexSim, RefPhSim, Div, \citet{Park_2019_ICCV}: LexSim, RefPhSim, \citet{Huang_2019_ICCV}: LexSim, RefPhSim, \citet{Laina_2019_ICCV}: LexSim, RefPhSim, \citet{Ke_2019_ICCV}: LexSim, RefPhSim, \citet{He_2019_ICCV}: LexSim, \citet{Vered_2019_ICCV}: LexSim, RefPhSim, Ret, \citet{Li_2019_ICCV}: LexSim, RefPhSim, \citet{Agrawal_2019_ICCV}: LexSim, RefPhSim, \citet{Gu_2019_ICCV}: LexSim, RefPhSim, \citet{Cornia_2019_CVPR}: LexSim, RefPhSim, \citet{Zheng_2019_CVPR}: LexSim, RefPhSim, \citet{Dognin_2019_CVPR}: LexSim, SenImSim, \citet{Feng_2019_CVPR}: LexSim, RefPhSim, \citet{Wang_2019_CVPR}: LexSim, RefPhSim, Div, \citet{Guo_2019_CVPR}: LexSim, Flu, \citet{Yin_2019_CVPR}: LexSim, \citet{Kim_2019_CVPR}: LexSim, Ret, \citet{Gao_2019_CVPR}: LexSim, RefPhSim, \citet{Qin_2019_CVPR}: LexSim, RefPhSim, \citet{Yang_2019_CVPR}: LexSim, RefPhSim, \citet{Deshpande_2019_CVPR}: LexSim, RefPhSim, Div, \citet{Biten_2019_CVPR}: LexSim, RefPhSim, \citet{Li_2019_CVPR}: LexSim, RefPhSim, \citet{Shuster_2019_CVPR}: LexSim, RefPhSim

\emph{Machine Learning} \citet{NEURIPS2019_680390c5}: LexSim, RefPhSim, \citet{NEURIPS2019_9c3b1830}: LexSim, RefPhSim, Div, \citet{NEURIPS2019_fecc3a37}: LexSim, RefPhSim, \citet{Wang_Chen_Hu_2019}: LexSim, RefPhSim, \citet{Song_Liu_Qian_Chen_2019}: LexSim, \citet{Li_Chen_Liu_2019}: LexSim, RefPhSim, Ret, \citet{Li_Jiang_Han_2019}: LexSim, \citet{Gao_Fan_Song_Liu_Xu_Shen_2019}: LexSim, RefPhSim, \citet{Chen_Mu_Xiao_Ye_Wu_Ju_2019}: LexSim, RefPhSim, \citet{Chen_Pan_Liu_Sun_2019}: LexSim, RefPhSim

\paragraph{2018.}
\emph{NLP} \citet{sharma-etal-2018-conceptual}: LexSim, RefPhSim, \citet{chen-etal-2018-attacking}: LexSim, \citet{madhyastha-etal-2018-end}: LexSim, RefPhSim, \citet{liu-etal-2018-simnet}: LexSim, RefPhSim, \citet{guo-etal-2018-improving}: LexSim, \citet{melas-kyriazi-etal-2018-training}: LexSim, \citet{lu-etal-2018-entity}: LexSim, Div, \citet{wang-etal-2018-object}: LexSim, RefPhSim, \citet{cohn-gordon-etal-2018-pragmatically}: Ret, \citet{chandrasekaran-etal-2018-punny}: NONE

\emph{Vision} \citet{Chen_2018_ECCV}: LexSim, \citet{Liu_2018_ECCV}: LexSim, RefPhSim, Ret, Div, \citet{Gu_2018_ECCV}: LexSim, Div, \citet{Dai_2018_ECCV}: LexSim, RefPhSim, \citet{Hendricks_2018_ECCV}: LexSim, Bias, \citet{Jiang_2018_ECCV}: LexSim, RefPhSim, \citet{Chen_2018_ECCV_b}: LexSim, \citet{Yao_2018_ECCV}: LexSim, RefPhSim, \citet{Chen_2018_CVPR}: LexSim, RefPhSim, \citet{Luo_2018_CVPR}: LexSim, RefPhSim, Ret, \citet{Mathews_2018_CVPR}: LexSim, RefPhSim, Flu, \citet{Chen_2018_CVPR_b}: LexSim, \citet{Aneja_2018_CVPR}: LexSim, RefPhSim, \citet{Anderson_2018_CVPR}: LexSim, RefPhSim

\emph{Machine Learning} \citet{NEURIPS2018_8bf1211f}: LexSim, RefPhSim, Div, \citet{NEURIPS2018_976abf49}: LexSim, RefPhSim, \citet{NEURIPS2018_d2ed45a5}: LexSim, RefPhSim, \citet{Jiang_Ma_Chen_Zhang_Liu_2018}: LexSim, Div, \citet{Chen_Ding_Zhao_Han_2018}: LexSim, \citet{Gu_Cai_Wang_Chen_2018}: LexSim, RefPhSim

\paragraph{2017.}
\emph{NLP} \citet{anderson-etal-2017-guided}: LexSim, RefPhSim

\emph{Vision} \citet{Shetty_2017_ICCV}: LexSim, RefPhSim, Div, \citet{Liu_2017_ICCV}: LexSim, \citet{Gu_2017_ICCV}: LexSim, RefPhSim, \citet{Pedersoli_2017_ICCV}: LexSim, \citet{Tavakoli_2017_ICCV}: LexSim, \citet{Yao_2017_ICCV}: LexSim, RefPhSim, \citet{Dai_2017_ICCV}: LexSim, RefPhSim, \citet{Vedantam_2017_CVPR}: LexSim, \citet{Gan_2017_CVPR}: LexSim, \citet{Ren_2017_CVPR}: LexSim, \citet{Lu_2017_CVPR}: LexSim, RefPhSim, \citet{Park_2017_CVPR}: LexSim, \citet{Yang_2017_CVPR}: NONE, \citet{Gan_2017_CVPR_b}: LexSim, \citet{Chen_2017_CVPR}: LexSim, \citet{Venugopalan_2017_CVPR}: LexSim, \citet{Yao_2017_CVPR}: LexSim, \citet{Sun_2017_CVPR}: LexSim, \citet{Rennie_2017_CVPR}: LexSim, \citet{Wang_2017_CVPR}: LexSim, RefPhSim, Div, \citet{Rohrbach_2017_CVPR}: LexSim

\emph{Machine Learning} \citet{NIPS2017_46922a08}: LexSim, Ret, \citet{NIPS2017_4b21cf96}: LexSim, RefPhSim, Div, \citet{Liu_Mao_Sha_Yuille_2017}: LexSim, \citet{Chen_Ding_Zhao_Chen_Liu_Han_2017}: LexSim, \citet{Li_Tang_Deng_Zhang_Tian_2017}: LexSim, \citet{Mun_Cho_Han_2017}: LexSim

\paragraph{2016.}
\emph{NLP} 

\emph{Vision} \citet{Hendricks_2016_CVPR}: LexSim, \citet{Johnson_2016_CVPR}: LexSim, Ret, \citet{You_2016_CVPR}: LexSim

\emph{Machine Learning} \citet{NIPS2016_9996535e}: LexSim, \citet{NIPS2016_eb86d510}: LexSim, Flu, \citet{yao2016empirical}: LexSim, \citet{Mathews_Xie_He_2016}: LexSim

\paragraph{2015.}
\emph{NLP} \citet{devlin-etal-2015-language}: LexSim, Flu, \citet{yagcioglu-etal-2015-distributed}: LexSim, \citet{chen-etal-2015-deja}: LexSim

\emph{Vision} \citet{Jia_2015_ICCV}: LexSim, \citet{Ushiku_2015_ICCV}: LexSim, \citet{Yu_2015_ICCV}: LexSim, \citet{Chen_2015_CVPR}: LexSim, Ret, Flu, \citet{Vinyals_2015_CVPR}: LexSim, Ret, \citet{Fang_2015_CVPR}: LexSim, Flu, \citet{Donahue_2015_CVPR}: LexSim, Ret, \citet{Karpathy_2015_CVPR}: LexSim, Ret

\emph{Machine Learning} \citet{pmlr-v37-xuc15}: LexSim, \citet{pmlr-v37-lebret15}: LexSim, \citet{mao2014deep}: LexSim, Ret, Flu

\paragraph{2014.}
\emph{NLP} \citet{mason-charniak-2014-nonparametric}: LexSim, \citet{mason-charniak-2014-domain}: LexSim, \citet{yatskar-etal-2014-see}: LexSim, \citet{kuznetsova-etal-2014-treetalk}: LexSim

\paragraph{2013.}
\emph{NLP} \citet{kuznetsova-etal-2013-generalizing}: LexSim, \citet{elliott-keller-2013-image}: LexSim

\paragraph{2012.}
\emph{NLP} \citet{kuznetsova-etal-2012-collective}: LexSim, \citet{mitchell-etal-2012-midge}: NONE

\paragraph{2011.}
\emph{NLP} \citet{li-etal-2011-composing}: LexSim

\emph{Machine Learning} \citet{NIPS2011_5dd9db5e}: LexSim

\paragraph{2010.}
\emph{NLP} \citet{feng-lapata-2010-many}: LexSim, \citet{aker-gaizauskas-2010-generating}: LexSim

\end{document}